\documentclass[letterpaper, 10 pt, conference]{ieeeconf}  

\IEEEoverridecommandlockouts                              

\usepackage{textcomp}
\usepackage{graphicx}
\usepackage{url}
\usepackage{amsmath} 
\usepackage{amssymb}  
\usepackage[procnumbered, ruled, linesnumbered]{algorithm2e}

\usepackage[frozencache,cachedir=.]{minted}
\usepackage{tcolorbox}
\tcbuselibrary{minted}
\PassOptionsToPackage{hyphens}{url}\usepackage{hyperref}

\newtcblisting{tcbpythoncode}{%
  colback         = yellow!5        ,
  colframe        = yellow!50!black ,
  listing only                      ,
  listing engine  = minted          ,
  minted language = python
}

\title{\LARGE \bf
Prototyping fast and agile motions for legged robots with Horizon
}

\author{Francesco Ruscelli\,$^{1}$, Arturo Laurenzi\,$^{1}$, Nikos G. Tsagarakis\,$^{1}$, and Enrico Mingo Hoffman$^{2}$
\thanks{$^{1}$Istituto Italiano di Tecnologia (IIT), Humanoids and Human Centered Mechatronics (HHCM), Via Morego 30, 16163 Genova, Italy}
\thanks{$^{2}$PAL Robotics, Carrer Pujades 77-79, 08005 Barcelona, Spain}%
}

\begin{document}

\maketitle
\thispagestyle{empty}
\pagestyle{empty}

\begin{abstract}
For legged robots to perform agile, highly dynamic and contact-rich motions, whole-body trajectories computation of under-actuated complex systems subject to non-linear dynamics is required.
In this work, we present hands-on applications of \emph{Horizon}, a novel open-source framework for trajectory optimization tailored to robotic systems, that provides a collection of tools to simplify dynamic motion generation.
%
Horizon was tested on a broad range of behaviours involving several robotic platforms:
we introduce its building blocks and describe the complete procedure to generate three complex motions using its intuitive and straightforward API.
\par \textit{Paper Type} -- Recent Work \cite{horizon}.
\end{abstract}
\section{INTRODUCTION}
Optimal control (OC) is a powerful tool for designing complex and dynamic motions for robotics platforms.
It seeks a trajectory that minimizes a performance index while satisfying a set of constraints to shape the resulting motion, obtaining the desired robot behaviour that guarantees feasibility and physics consistency.
Horizon is an intuitive tool based on CasADi \cite{Andersson2019} that uses OC to rapidly generate robot trajectories, providing the user with a pipeline encompassing the whole problem of loading a robot model, imposing the preferred behaviour, and obtaining a dynamically-feasible motion that satisfies the user's instructions.
It allows to set up and solve trajectory optimization (TO) problems in both an offline and receding horizon fashion, seeking to simplify the pipeline of optimal motion planning for robotic systems.
Horizon is generic enough to include all the necessary tools to prototype a dynamic manoeuvre, co-design a robot structure, or produce periodic gaits.

\section{BUILDING BLOCKS}\label{sec:ingredients}
The process of prototyping robot motions follows a well-defined structure that can be adopted for different goals and robots. This Section offers descriptions and API examples of the main blocks required to generate a dynamic motion by constructing and solving an OC problem using Horizon.
\subsection{Model acquisition}\label{subsec:model_acquisition}
The parser module retrieves the model so as to be compatible with the underlying symbolic framework.
It provides the robot dynamics in a symbolic form, so that it can be injected into the optimization: throughout the problem definition, model information can be obtained as symbolic variables that can be used in constraints and objective functions. Leveraging Pinocchio~\cite{carpentier2019pinocchio}, it offers useful robotics-oriented algorithms, such as direct and inverse kinematics and dynamics.
The \textit{HorizonParser} guarantees ease-of-use and compatibility, as it supports the Universal Robot Description Format (URDF).
\begin{tcbpythoncode}
urdf = open(urdf_file, 'r').read()
model = CasadiKinDyn(urdf)
\end{tcbpythoncode}
\begin{figure}[!t]
    \centering
    \includegraphics[width=0.59\columnwidth]{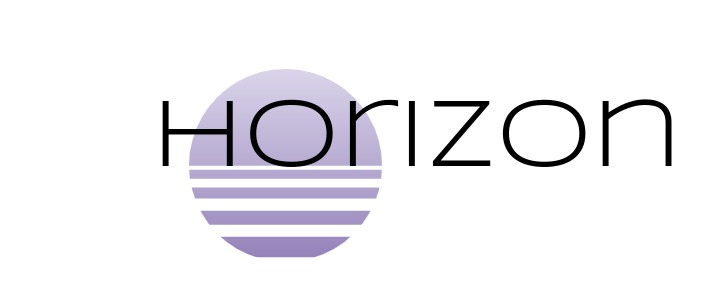}
    \caption{The Horizon framework is available at: \protect\url{https://github.com/ADVRHumanoids/horizon}}
    \label{fig:horizon_logo}
\end{figure}
\subsection{Horizon specification}\label{subsec:horizon_specification}
The TO problem can be set up as a fixed horizon, to design non-periodic manoeuvres with a final goal, or in a receding horizon fashion, to generate periodic motion or continuous movements. Horizon allows both scenarios, given a suitable number of nodes to discretize the time window of the problem. This choice implies a trade-off between the resolution of the time grid and the problem dimension.
\begin{tcbpythoncode}
prb = Problem(n_nodes)
\end{tcbpythoncode}
\subsection{Transcription selection}\label{subsec:transcription_selection}
Transcription methods are used for transcribing the OC problem into a non-linear programming (NLP) one, which can be efficiently solved by state-of-the-art solvers. The input and state trajectories, parametrized as decision variables over the horizon, are connected by junction nodes and gap constraints to enforce dynamic feasibility.
The user can choose among two available built-in strategies: multiple shooting and direct collocation. Once specified, Horizon formalizes it as a set of constraints on each node to guarantee continuity.
\begin{tcbpythoncode}
tr = 'multiple_shooting' #'collocation'
int_type = 'RK4' # 'leapfrog', 'euler'
Transcriptor(tr, prb, int_type)
\end{tcbpythoncode}
\begin{figure*}[ht]
    \centering
    \includegraphics[width=\textwidth]{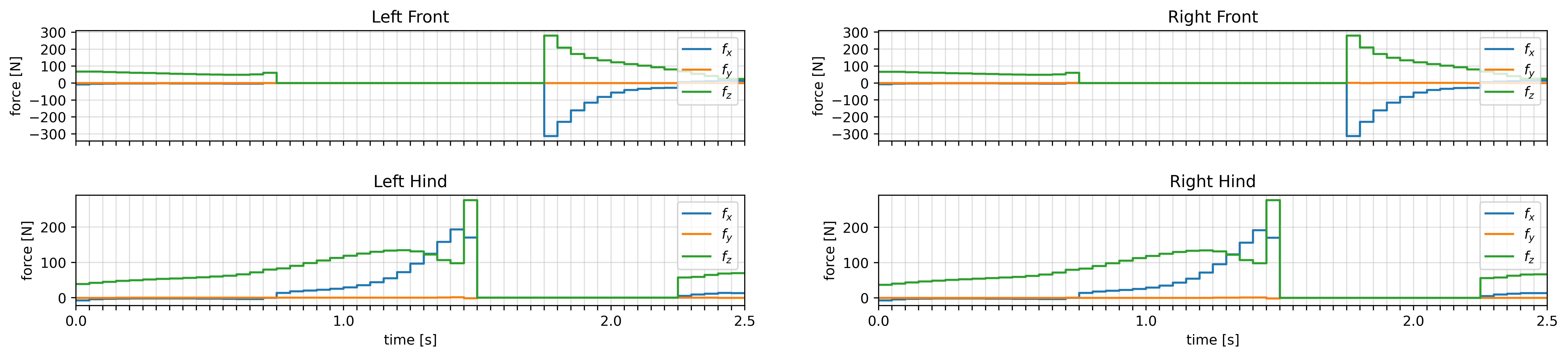}
    \caption{Time histories of the contact forces during the leap. Notice how the robot jumps with the hinds leg and lands using the forward legs.}
    \label{fig:leap_force}
\end{figure*}
\subsection{System formulation}\label{subsec:system_formulation}
While its core functionalities can be used even outside the robotics scope, Horizon's main target is robotic applications, as it provides specific built-in methods and robotics-oriented utilities. Therefore, the system taken into consideration in this work is the generic floating-base robotic system:
\begin{equation}\label{eq:floating_base_dynamics}
     \boldsymbol{M}(\boldsymbol{q})\dot{\boldsymbol{\nu}}+\boldsymbol{h}(\boldsymbol{q},\boldsymbol{\nu}) = \mathbf{S}\boldsymbol{\tau} + \mathbf{J}_c^T(\boldsymbol{q})\boldsymbol{f}_{c}
\end{equation}
which describe its dynamics with the usual symbols. 
The optimization problem is constructed from the generalized coordinates 
$\boldsymbol{q}$ 
and velocities 
$\boldsymbol{\nu}$ 
of the floating-base system:
\begin{equation}
    \boldsymbol{q} = \begin{bmatrix}\boldsymbol{p}\\ \boldsymbol{\rho}\\ \boldsymbol{\theta} \end{bmatrix}, \quad
    \boldsymbol{\nu} = \begin{bmatrix}\dot{\boldsymbol{p}}\\ \boldsymbol{\omega}\\ \dot{\boldsymbol{\theta}} \end{bmatrix}
\end{equation}
where 
$\boldsymbol{p}, \boldsymbol{\rho}, \boldsymbol{\theta}$ are the linear position, the orientation of the floating-base and the vector of joint positions respectively. Consequently, $\dot{\boldsymbol{p}}, \boldsymbol{\omega}, \dot{\boldsymbol{\theta}}$ are the linear and angular velocity of the floating base and the vector of joint velocities.
\begin{tcbpythoncode}
q = prb.createStateVar('q', q_dim)
v = prb.createStateVar('v', v_dim)
\end{tcbpythoncode}

\begin{figure}[!t]
    \centering
    \includegraphics[width=1\columnwidth]{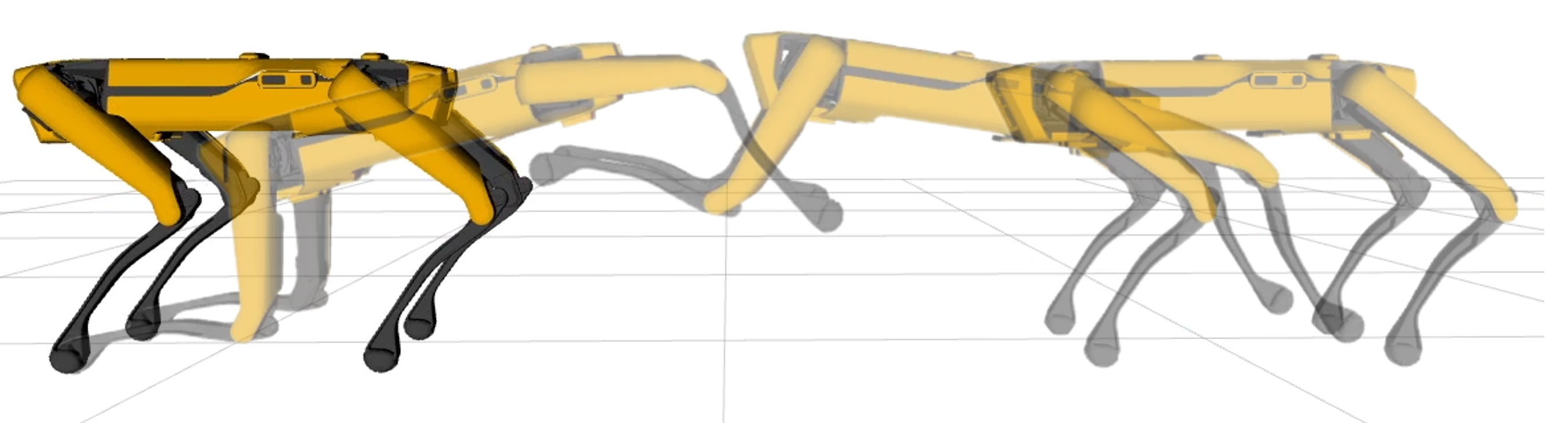}
    \caption{Frame sequence of Spot\textsuperscript \textregistered{} while performing a leaping motion. The quadruped robot jumps using the hind legs to provide thrust, and finally lands on the front legs at a specified position.}
    \label{fig:spot_leap}
\end{figure}
\subsection{Feasibility injection}\label{subsec:feasibility_injection}
A necessary ingredient for floating-base systems is the set of constraints to impose feasibility, i.e. the system under-actuation.
The equations of motion are provided by the model (see Subsection \ref{subsec:model_acquisition}). Given the vector of Cartesian forces at each contact,
the inverse dynamics torques can be retrieved from the general form of \eqref{eq:floating_base_dynamics}.
The first six elements, corresponding to the effort on the virtual joints $\boldsymbol{\tau}_{fb}$, are constrained to zero, to enforce under-actuation.
\begin{tcbpythoncode}
i_d = model.InverseDynamics(contacts)
fb_tau = i_d.call(q, v, a)[:6]
prb.createConstr('feas', fb_tau)
\end{tcbpythoncode}
\begin{figure*}
    \centering
    \includegraphics[width=0.9\textwidth]{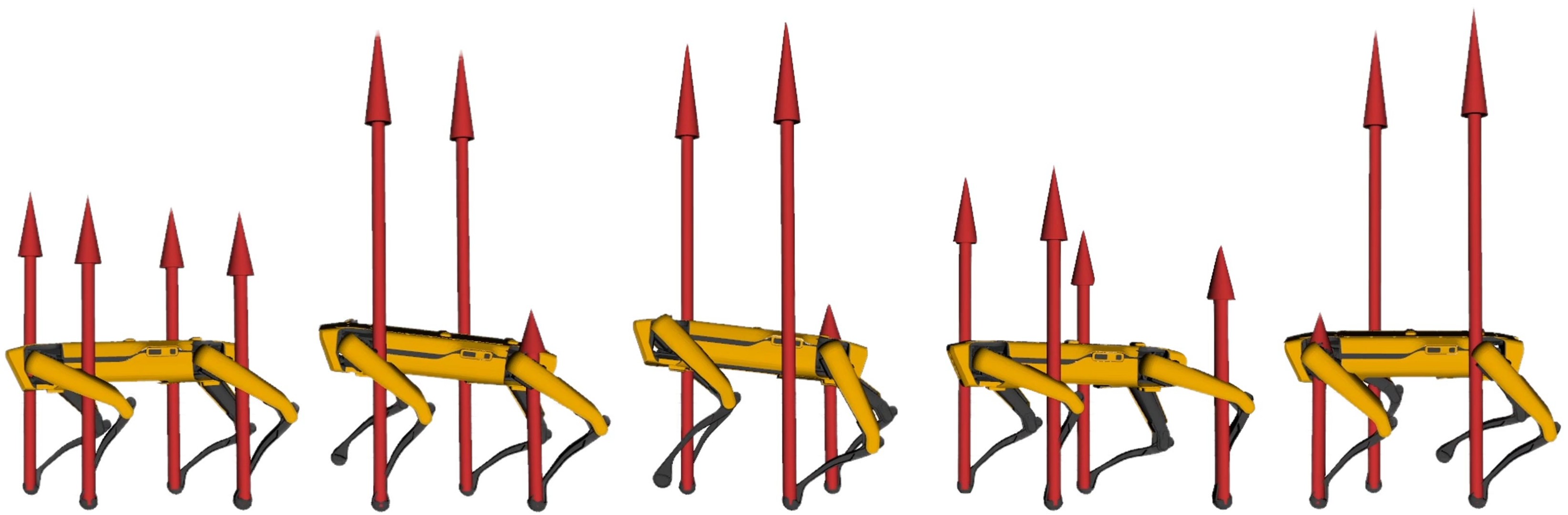}
    \caption{Frame sequence of Spot\textsuperscript \textregistered \ walking in a receding horizon fashion.}
    \label{fig:spot_walk}
\end{figure*}
\subsection{Contact schedule}\label{subsec:contact_schedule}
A generic manoeuvre consists of a sequence of contacts of the robot with the environment that are established and broken. In this work, contact scheduling is enforced as follows: when the contact is established, the Cartesian velocity at the end-effector frame $\mathbf{\dot{p}}_c$ is constrained to be zero and the contact forces $\boldsymbol{f}_c$ greater than zero.
\begin{tcbpythoncode}
# contact on, n_st = [stance_nodes]
DFK = model.frameVelocity('contact')
v_c = DFK(q, q_dot)['ee_vel_linear']
prb.createConstr('v_c', v_c, n_st)
prb.createConstr('f_c', f_c, n_st, upper_bounds=inf)
\end{tcbpythoncode}
\noindent
Conversely, breaking the contact is encoded by constraining the Cartesian forces at the end-effector to be zero.
The constrains presented above can be combined along the horizon to assign phases of the robot motion to specific portions of the trajectory, ultimately shaping the resulting motion.
\begin{tcbpythoncode}
# contact off, n_sw = [swing_nodes]
prb.createConst('f_c', f_c, n_sw)
\end{tcbpythoncode}
\subsection{Solver selection}\label{subsec:solver_selection}
According to the authors' experience, and given the wide variety of NLP problems that come out of an OC problem, it is very likely that a one-size-fits-all solver does not exist. On the contrary, a specific solver can be more prone to finding a suitable solution to the task at hand. Horizon makes several different solvers available to the user, allowing to switch easily between them:
\begin{enumerate}
    \item A custom implementation of a sparse, Gauss-Newton Sequential Quadratic Programming (GN-SQP) algorithm;
    \item A custom implementation of a multiple-shooting Iterative Linear-Quadratic Regulator (ILQR);
    \item IPOPT \cite{ipopt} and similar state-of-the-art solvers.
\end{enumerate}
\begin{tcbpythoncode}
slv_type = 'ipopt' # 'ilqr', 'gnsqp'
solver = Solver(slv_type, prb, opts)
\end{tcbpythoncode}
\section{PROTOTYPING A MOTION}
The building blocks presented in Section \ref{sec:ingredients} can be assembled to generate a desired dynamic motion of a robot. Horizon allows to combine them seamlessly in a simple and intuitive way: this Section describes how three motions are generated, drawing parallels between the theoretical design of the OC problems and their implementation in Horizon.
\subsection{Common features}\label{subsec:common_features}
Two features are common to the robotic applications presented in this section: first, the formulation of the dynamical system:
we choose to assemble the state vector of the system as joint position and velocity combined, whereas the input is chosen as the vector of generalized accelerations and forces:
\begin{equation}\label{eq:state_and_input}
     \mathbf{x}  = \begin{bmatrix} \boldsymbol{q} \\ \boldsymbol{\nu} \end{bmatrix}, \quad
     \mathbf{u} = \begin{bmatrix} \mathbf{\dot{\boldsymbol{\nu}}} \\ \boldsymbol{f}_c \end{bmatrix}
\end{equation}
\begin{tcbpythoncode}
# acc dim: n_v # f dim: n_f
a = prb.createInputVar('a', n_v)
f_c = prb.createInputVar('f_c', n_f)
\end{tcbpythoncode}
\noindent
Given the state and the input as decision variables, Horizon allows to inject the system dynamics in the problem:
\begin{tcbpythoncode}
xdot = getDynamics()
prb.setDynamics(xdot)
prb.setDt(dt)
\end{tcbpythoncode}
\noindent
The second common feature consists of the TO initial conditions: on the first node the joint position is set to the initial robot configuration while the velocity is bound to zero.
\begin{tcbpythoncode}
# initial bounds (at node 0)
q.setBounds(lb=q0, ub=q0, n=0)
v.setBounds(lb=0, ub=0, n=0)
\end{tcbpythoncode}
\noindent
Finally, regularization terms are introduced for the input: both the velocity $\boldsymbol{\nu}$ and the contact forces $\boldsymbol{f}_c$ are minimized.
\begin{tcbpythoncode}
prb.setCost('min_f_c', f_c)
prb.setCost('min_v', v)
\end{tcbpythoncode}
\noindent
\subsection{Leaping}\label{subsec:leaping}
\begin{figure*}[t]
    \centering
    \includegraphics[width=0.8\textwidth]{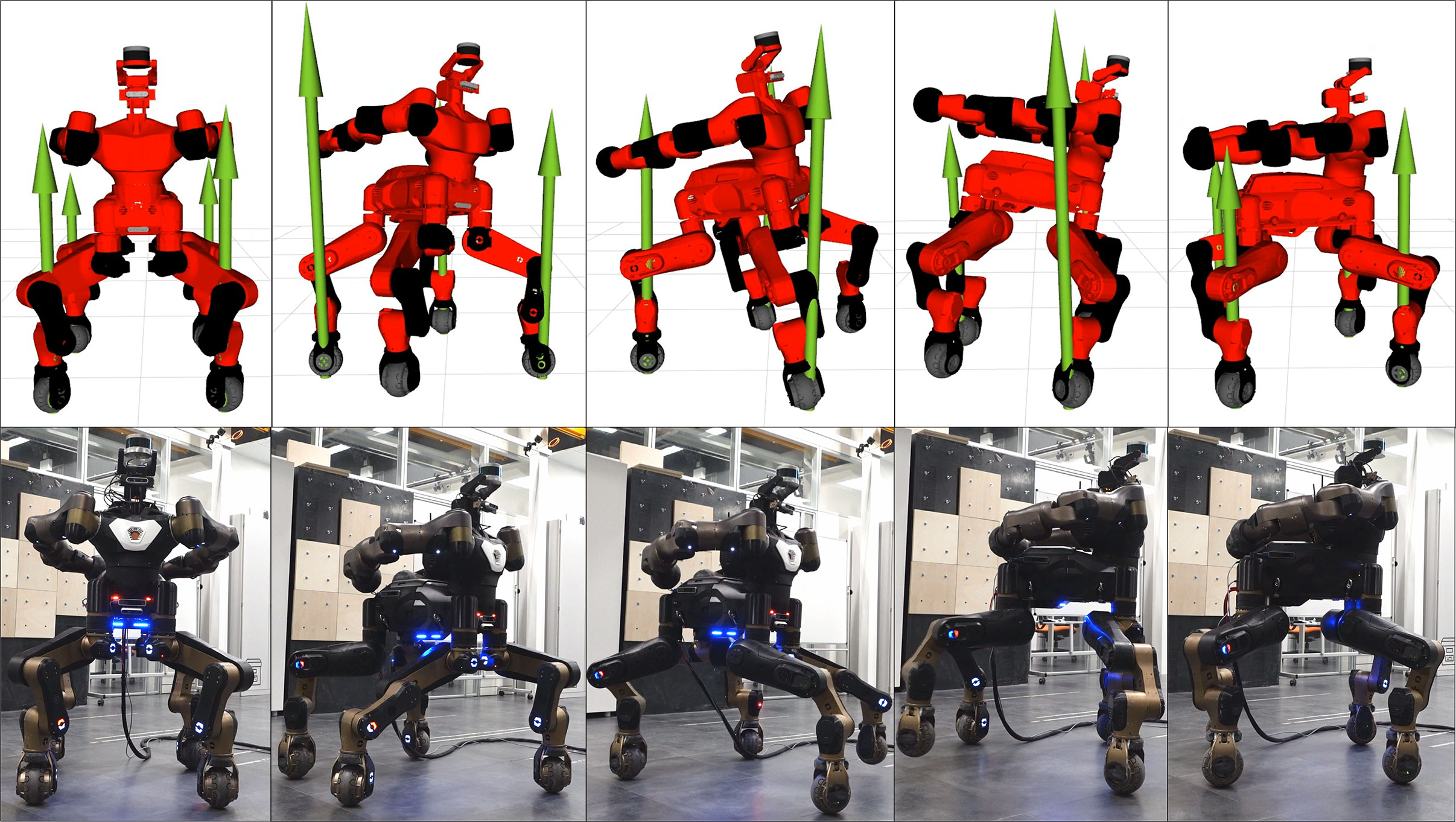}
    \caption{Frame sequence of Centauro while performing a 90 deg in-place turn. The green arrows represent the contact forces at each foot.}
    \label{fig:centauro_twist}
\end{figure*}
The first optimization problem ought to find joint acceleration, torques and contact forces that correspond to a leap motion of a quadruped robot.
The model acquired (see \ref{subsec:model_acquisition}) is Spot\textsuperscript\textregistered. The finite-horizon (see \ref{subsec:horizon_specification}) is selected, as the motion is aperiodic. Generally, a finite-horizon trajectory requires terminal constraints: we impose the zero velocity at the end of the motion, as the robot should be standing still. This rule amounts to adding a single constraint at the last node forcing the joint velocity to be zero:
\begin{tcbpythoncode}
prb.setFinalConstr('v_final', v)
\end{tcbpythoncode}
\noindent
The transcription method in \ref{subsec:transcription_selection} is chosen as the multiple shooting. The system is formulated as in \ref{subsec:system_formulation} and \ref{subsec:feasibility_injection}. The contact schedule (see \ref{subsec:contact_schedule}) to enforce a leaping motion is chosen (Figure \ref{fig:leap_force} shows the contact pattern as the contact forces are constrained to zero). Finally, we decide upon IPOPT, the large scale non-linear solver.
To obtain the forward motion in the jump, a constraint is added for the floating base position $\boldsymbol{p}$ at the last node. Additionally, a postural for the robot is set to ensure a valid final pose:
\begin{equation}
\boldsymbol{p}^N = \boldsymbol{p}_{\text{des}},  \quad
\boldsymbol{\rho}^N = \boldsymbol{\rho}^0, \quad
\boldsymbol{\theta}^N = \boldsymbol{\theta}^0
\end{equation}
\begin{tcbpythoncode}
prb.setFinalConstr(x[:3] - fbp_des)
prb.setFinalConstr(x[3:6] - fbr0)
prb.setFinalConstr(x[7:] - q0)
\end{tcbpythoncode}
\subsection{Crawling}\label{crawling90}
The goal for this optimization is prototyping a 90 deg turn of the robot CENTAURO within two full gait cycles (8 steps). The robot URDF, the system formulation presented in Subsection \ref{subsec:common_features}, the finite-horizon (\ref{subsec:horizon_specification}), the multiple shooting transcription method (\ref{subsec:transcription_selection}) and the contact scheduling (\ref{subsec:contact_schedule}) are selected.
To obtain the twisting motion during the jump, it is enough to constrain the robot postural and the floating base at the last node. Specifically, the orientation of the floating base is set to the desired one, while the remaining elements, i.e. the floating base $\boldsymbol{p}$ and the joint $\boldsymbol{\theta}$ positions, are constrained to the initial configuration:
\begin{equation}
\boldsymbol{p}^N = \boldsymbol{p}^0, \quad
\boldsymbol{\rho}^N = \boldsymbol{\rho}^{\text{des}}, \quad
\boldsymbol{\theta}^N = \boldsymbol{\theta}^0
\end{equation}
where $\boldsymbol{\rho}_\text{des}$ is the quaternion corresponding to a rotation of 90 deg.
Finally, a so-called \textit{clearance constraint} imposes, for each foot, a fixed swing trajectory on the $z$-axis.
\begin{tcbpythoncode}
FK = model.fk(frame)
p_c = FK(q)['ee_pos_linear']
prb.setConstr('clear', p_c[2] - z_sw)
\end{tcbpythoncode}
\subsection{Walking}\label{walking}
This application shows how Horizon can be used in a Horizon receding horizon scenario, i.e. a continuous walking motion of Spot\textsuperscript\textregistered \ (see Figure \ref{fig:spot_walk}). 
Once again, the common formulation of the problem presented in Subsection \ref{subsec:common_features} is used.
However, the horizon (\ref{subsec:horizon_specification}) is chosen to be receding. This implies some design choices: the contact schedule presented in Subsection \ref{subsec:contact_schedule} is defined for one cycle only and fed to Horizon. 
Then, instead of imposing final constraints, a constant velocity reference is set as a cost function for the base link of the robot. 
\begin{tcbpythoncode}
prb.createCost('v_ref', v - v_ref)
\end{tcbpythoncode}
This translates into the optimizer searching for a periodic gait given the contact schedule.
Finally, a postural cost is added during the whole motion to keep the robot configuration, while walking, as close as possible to the initial pose.
\begin{tcbpythoncode}
prb.createCost('postural', q[7:] - q0)
\end{tcbpythoncode}
The resulting problem was solved online with the custom ILQR solver.
\section{CONCLUSIONS}
Horizon was successfully tested on several robots\footnote{Clips of more applications are gathered at: \url{https://youtube.com/playlist?list=PL7c1ZKncPan6ngCiYJHP0uciQygY1hl5I}.} to generate a vast range of behaviours, proving to be a suitable tool for prototyping motions. 
Horizon aims at simplifying the process of generating agile motions of robotic platforms without losing the performance that state-of-the-art algorithms can offer. 
While Horizon is mature enough for complex applications, this project is still ongoing: the next milestone will extend the pipeline with a motion control layer to open the path for MPC applications, bridging the gap between simulation and real robot deployment.

\bibliographystyle{IEEEtran}	
\bibliography{bib}
\end{document}